\documentclass[10pt,twocolumn,letterpaper]{article}

\usepackage[pagenumbers]{wacv}
\usepackage[skip=5pt]{caption}
\usepackage{times}
\usepackage{epsfig}
\usepackage{graphicx}
\usepackage{amsmath}
\usepackage{amssymb}
\usepackage{booktabs}
\usepackage{multirow}
\usepackage{float}
\usepackage{subcaption}
\usepackage{hyperref}
\usepackage[accsupp]{axessibility} 

\begin{document}
\title{FastPose-ViT: A Vision Transformer for Real-Time Spacecraft Pose Estimation}

\author{
Pierre Ancey$^{1}$ \quad
Andrew Price$^{1}$ \quad
Saqib Javed$^{1}$ \quad
Mathieu Salzmann$^{1,2}$ \\
[0.2cm]
$^{1}$EPFL \quad
$^{2}$Swiss Data Science Center \\
Lausanne, Switzerland \\
[0.1cm]
{\tt\small \{firstname.lastname\}@epfl.ch}
}

\twocolumn[{
\renewcommand\twocolumn[1][]{#1}
\maketitle
\begin{center}
    \includegraphics[width=1.0\linewidth]{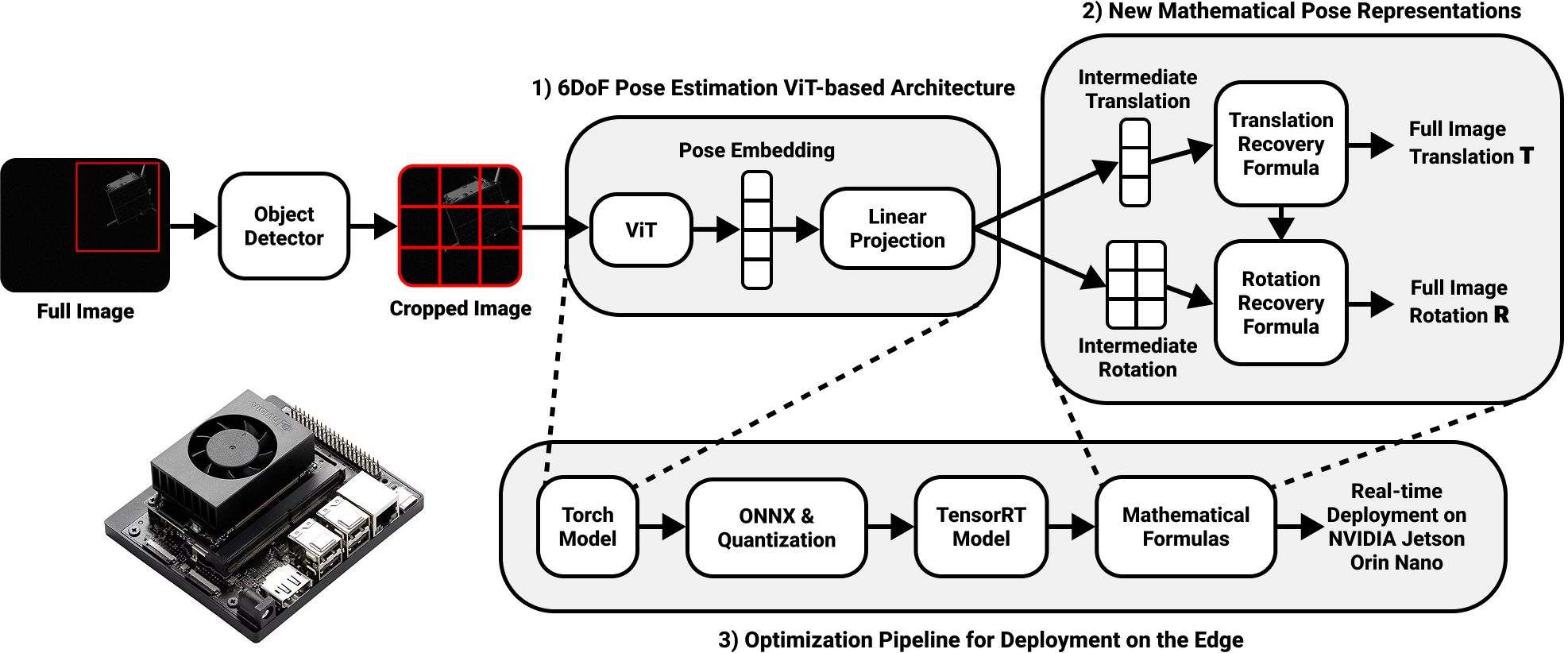}
    \vspace{2pt}
    \captionof{figure}{\textbf{Overview of the FastPose-ViT pipeline.} A bounding-box detector first extracts the spacecraft from the full image, producing a cropped input for the pose network. Then, (1) a ViT-based architecture regresses intermediate pose parameters from the crop; (2) geometric recovery formulas convert these into the full-frame translation and rotation; (3) the trained model is exported, quantized, and optimized for real-time deployment on a NVIDIA Jetson Orin Nano.}
    \label{fig:overview}
\end{center}
\vspace{12pt} 
}]

\begin{abstract}
   \textit{Estimating the 6-degrees-of-freedom (6DoF) pose of a spacecraft from a single image is critical for autonomous operations like in-orbit servicing and space debris removal. Existing state-of-the-art methods often rely on iterative Perspective-n-Point (PnP)-based algorithms, which are computationally intensive and ill-suited for real-time deployment on resource-constrained edge devices. To overcome these limitations, we propose FastPose-ViT, a Vision Transformer (ViT)-based architecture that directly regresses the 6DoF pose. Our approach processes cropped images from object bounding boxes and introduces a novel mathematical formalism to map these localized predictions back to the full-image scale. This formalism is derived from the principles of projective geometry and the concept of ``apparent rotation", where the model predicts an apparent rotation matrix that is then corrected to find the true orientation. We demonstrate that our method outperforms other non-PnP strategies and achieves performance competitive with state-of-the-art PnP-based techniques on the SPEED dataset. Furthermore, we validate our model’s suitability for real-world space missions by quantizing it and deploying it on power-constrained edge hardware. On the NVIDIA Jetson Orin Nano, our end-to-end pipeline achieves a latency of ~75 ms per frame under sequential execution, and a non-blocking throughput of up to 33 FPS when stages are scheduled concurrently.
   }
\end{abstract}

\newpage
\section{Introduction}

The increasing complexity of space missions has created a pressing need for autonomous systems capable of performing tasks such as in-orbit servicing, active debris removal, and asteroid mining. A fundamental requirement for these operations is the precise and rapid estimation of a spacecraft's 6-degrees-of-freedom (6DoF) pose --its translation and attitude-- from a single monocular image.

While state-of-the-art approaches have achieved high accuracy using hybrid pipelines that combine keypoint detection with iterative Perspective-n-Point (PnP) solvers~\cite{SPEEDPark2023, Wang22, Gerard2019, huo2020fast, lotti2023deep}, these methods are inherently slow and poorly suited to real-time deployment on edge computing platforms due to their iterative nature. Direct regression, in which a neural network predicts 6DoF pose in a single feed-forward pass~\cite{Proenca2019, Musallam2022}, offers a promising alternative due to its computational efficiency. However, existing direct regression methods have typically underperformed in accuracy relative to their hybrid counterparts~\cite{Kisantal20, Cui2023}.

To address this limitation, we propose a new architecture based on the Vision Transformer~\cite{dosovitskiy2020transformers} (ViT) that aims to bridge the gap between speed and accuracy in spacecraft pose estimation.
Our primary contributions are: 

\begin{itemize}
    \item A ViT~\cite{dosovitskiy2020transformers}-based architecture designed specifically for efficient and direct 6DoF pose regression.
    \item A reformulation of the regression targets based on geometric principles. Our model predicts an apparent rotation; we then apply a closed-form correction, derived from projective geometry, to recover the true rotation. 
    \item A pipeline that exports the model from PyTorch to TensorRT, enabling FP16 quantization without accuracy loss and real-time inference on edge devices. 
\end{itemize}

We evaluate our method on the \textsc{SPEED}~\cite{Sharma19} and \textsc{SPEED+}~\cite{Park2021} datasets, where it sets a new state-of-the-art for non-PnP-based methods. Crucially, our approach achieves performance that is competitive with top-tier PnP-based methods on \textsc{SPEED} while also operating in real-time on a small edge device, demonstrating that direct regression is a capable alternative for real-world spacecraft applications.

\section{Related Work}

\paragraph{Spacecraft Pose Estimation.}
Spacecraft pose estimation involves unique challenges such as high specularity, extreme dynamic ranges, and non-diffuse lighting. This has been presented in different settings including SwissCube~\cite{wdrposehu2021}, URSO~\cite{Proenca2019}, SPARK~\cite{spark2022}, Synthetic-Minerva~\cite{Price21} and PocketQube~\cite{Sajjad2025}. However, the de facto benchmarks for monocular spacecraft 6DoF pose estimation are the European Space Agency's (ESA) Satellite Pose Estimation Challenges~\cite{ESALeaderboard, ESALeaderboard2}, which introduced the SPEED and SPEED+ datasets now commonly used in the space community. Against these benchmarks, many 6DoF pose estimation strategies have been tested, and can broadly be categorized as hybrid or direct~\cite{Cui2023}.

\textbf{Hybrid methods} combine deep learning with geometric solvers~\cite{Gerard2019, Chen2019, wdrposehu2021}. Top performers on leaderboards~\cite{ESALeaderboard, ESALeaderboard2}, such as UniAdelaide~\cite{Chen2019} and CVLab~\cite{wdrposehu2021}, detect 2D keypoints on the object and then use a PnP solver with RANSAC~\cite{Fischler1981} to compute the pose. Li et al.~\cite{Li2022}, predict keypoints along with uncertainty values to guide the PnP algorithm, improving robustness against outliers.
While accurate, these methods are computationally intensive. For example, one PnP-based approach~\cite{Wang22} reported \textbf{150~ms per image ($\sim$7~FPS)} on a high-performance NVIDIA RTX 2080Ti GPU (peak draw 250 Watts) and an i7-7700K CPU, far exceeding the typical power budgets of a spacecraft ($\sim$1–25 Watts). This high latency stems from the multi-stage pipelines and iterative solvers these methods rely on.

\textbf{Direct methods} bypass intermediate steps by regressing pose parameters directly from the image. URSONet~\cite{Proenca2019} uses a CNN to regress translation and classify orientation into discrete bins. E-PoseNet~\cite{Musallam2022} introduces equivariant CNNs to improve robustness to geometric transformations in a lightweight architecture. However, these direct methods still fall short of the accuracy achieved by hybrid approaches. To bridge this gap while preserving efficiency, we extend the direct regression paradigm by introducing geometrically derived regression targets and replacing conventional CNNs with a ViT-based architecture. 

\paragraph{Vision Transformers for Pose Estimation.}
The application of Vision Transformers (ViTs) to direct 6DoF pose estimation has been explored in PViT-6D~\cite{Stapf2023}, which employs a Multiscale ViT~\cite{Fan21} and was evaluated on the BOP benchmark~\cite{hodan2018bop} for terrestrial objects. While conceptually interesting, PViT-6D suffers from practical limitations: the architecture is complex, pretrained weights are unavailable, and the implementation is difficult to reproduce. In contrast, FoundationPose~\cite{foundationposewen2024} demonstrates that ViTs can achieve high precision, but its multi-stage pipeline is not designed for deployment on edge devices. These limitations motivated us to design a custom ViT-based architecture that is both lightweight and reproducible, providing a practical solution for direct pose estimation in space applications.

\paragraph{Rotation Representations.}
The representation of rotation is a critical aspect of 6DoF pose estimation. Common representations such as quaternions and Euler angles suffer from ambiguity and discontinuity issues, respectively. To address this, Zhou et al.~\cite{Zhou19} proposed continuous 6D representations to create a more stable learning target, showing improved performance. While our work also leverages a continuous 6D representation for rotation matrices, our key innovation lies in reformulating the regression target itself.

\section{Methodology}

Our goal is to predict the translation vector $T \in \mathbb{R}^3$ and rotation matrix $R \in SO(3)$ of a spacecraft relative to the camera frame from a single RGB image. Our method, FastPose-ViT, is built upon a ViT backbone and leverages bounding box priors to improve efficiency and accuracy.

\subsection{Vision Transformer Architecture}
The standard ViT architecture~\cite{dosovitskiy2020transformers} was designed for image classification on large datasets like ImageNet~\cite{deng2009imagenet}. It splits an image into non-overlapping patches, linearly projects the patches into embeddings called “tokens,” and prepends a learnable [CLS] token to aggregate global information via self-attention for the final classification task.

While our goal is 6DoF regression rather than classification, we can adapt this architecture through transfer learning, leveraging the strong spatial representations acquired during pre-training. For our pose regression task, we modify the architecture as follows:

\begin{enumerate}
    \item The input image is first cropped using a bounding box obtained from an object detector, focusing the ViT's attention on relevant patches.
    \item The original classification Multi-Layer Perceptron (MLP) is discarded.
    \item The final `[CLS]' token embedding is passed through a single linear projection layer to predict translation and rotation parameters.
\end{enumerate}

We also experimented with deeper MLPs as regression heads, but found no consistent improvement over the simple linear projection. This suggests that the ViT backbone already encodes sufficient global information alone.

To find the best trade-off between accuracy and computational cost, 
We evaluate ViT-B backbones, where ViT-B refers to the Base variant of the Vision Transformer architecture introduced by Dosovitskiy et al.~\cite{dosovitskiy2020transformers},
at two input resolutions; the models are summarized in Table~\ref{table:vit_metrics_updated}. Additional experiments with slower and less performative ViT backbones are shared in the supplementary material.

\begin{table}[b]
  \centering
  {\small{
  \begin{tabular}{@{}l r r l c@{}}
    \toprule
    Backend & Params (M) & GFLOPS & Image Size & Patch \\
    \midrule
    ViT-B-224/16 & 85.8 & 17.5 & 224x224 & 16x16 \\
    ViT-B-384/16 & 86.1 & 55.5 & 384x384 & 16x16 \\
    \bottomrule
  \end{tabular}
  }}
  \caption{Characteristics of the ViT-B/16 backbones with linear projection layer evaluated in this work.}
  \label{table:vit_metrics_updated}
\end{table}

\subsection{Data Augmentations}
To improve model robustness and prevent overfitting, we employ extensive data augmentation strategies at both the pixel and spatial levels.

\paragraph{Pixel-Level Augmentations.}
To improve robustness to the challenging imaging conditions encountered in orbit, we apply a diverse set of pixel-level transformations using the Albumentations library~\cite{Albumentations}. These include variations in illumination, lens and optical effects, sensor noise, motion blur, and compression artifacts, each applied stochastically with probability $0.5$. All images are then normalized to ImageNet statistics and converted to tensors. This pipeline replicates the lighting, optical, and sensor degradations typical of spacecraft imagery, encouraging the model to learn invariant representations.

\paragraph{Bounding-Box Augmentations.}
To avoid overfitting to perfectly framed crops, which is unrealistic in practice, we deliberately perturb the bounding boxes used to generate input crops. This forces the model to rely on actual pose cues rather than memorizing canonical crop sizes and positions.

To obtain the initial bounding boxes, we hand-annotated all images in the \textsc{SPEED} training set and trained an LW-DETR~\cite{lwdetr} detector on these annotations, modifying its output head to predict a single spacecraft class instead of COCO’s 80 classes~\cite{lin2014microsoft}. We then applied this detector to produce bounding boxes for the \textsc{SPEED} evaluation set and for all splits of \textsc{SPEED+}. These automatically generated boxes serve as the baseline from which augmentations are applied.

During training of the pose network, each box $B=(x_{\min},y_{\min},x_{\max},y_{\max})$ is randomly expanded or contracted along each side by up to $10\%$ of its width or height. The resulting box is clipped to image boundaries and discarded if it becomes invalid (e.g., degenerate or too small). This strategy increases robustness to the noisy detections commonly encountered in real settings.

\paragraph{Spatial-Level Augmentations.} We apply in-plane rotations to the images to augment the dataset with varied viewpoints. A key challenge is that rotating the image also changes the ground-truth pose. For an in-plane rotation of angle $\theta$ around the image center, the new rotation matrix $R_{new}$ and translation vector $T_{new}$ can be computed from the original pose $(R, T)$ and the intrinsic matrix $K$ as:
\begin{equation}
    R_{new} = K^{-1}M(\theta) K R, \quad T_{new} = K^{-1}M(\theta) K T,
\end{equation}
where $M(\theta)$ is the 2D rotation matrix in homogeneous coordinates. The term $K^{-1}M(\theta) K$ simplifies to a pure 3D rotation matrix $R_z(\theta)$ around the camera's Z-axis (the axis orthogonal to the image plane). For completeness, the spatial augmentation derivation is shared in the supplementary material. The pose is updated as
\begin{equation}
    R_{new} = R_z(\theta)R, \quad T_{new} = R_z(\theta)T.
\end{equation}
This process allows for the accurate relabeling of our spatially augmented data. During training, a spatial augmentation is applied with a 50\% probability, where one of two rotation ranges is selected with equal likelihood: Either a minor rotation within $[-20^{\circ}, 20^{\circ}]$ or a major rotation within $[160^{\circ}, 200^{\circ}]$.

These specific ranges were chosen because they guarantee that, for the extreme majority of SPEED and SPEED+ samples, at least part of the satellite remains visible in the augmented crop, ensuring that a valid bounding box can still be defined.

\subsection{Geometric Reformulation of Pose Regression for Cropped Images}
\label{sec:geom_reform}

A key innovation of our work is reformulating the pose regression targets to align with the localized visual information available in a cropped image. While using crops is highly efficient for ViTs, it presents a challenge; the full image absolute pose must be recovered from the cropped image apparent pose.

Our solution is to train the network to predict a set of intermediate, normalized parameters that are directly observable in the crop, rather than the absolute pose itself. We then use a geometric post-processing step to deterministically recover the full-frame pose from these intermediate values. We address the translation and rotation components separately.

\subsubsection{Translation via Normalized Coordinates}

\paragraph{Step 1: Formulating a Normalized Depth Target ($U_z$).}
The act of cropping and resizing an image patch is analogous to a digital zoom. We relate the object's true depth $Z$ to a normalized ``crop-depth proxy", $U_z$, using the scaling factors $s_x$ and $s_y$ derived from the crop and full-image dimensions as
\begin{equation}
s_x := \frac{W}{\alpha\,w},\qquad
s_y := \frac{H}{h},
\end{equation}
where $(W,H)$ and $(w,h)$ are the full-image and crop dimensions, respectively, and $\alpha$ is a dataset-specific aspect-ratio factor (1.6 for \textsc{SPEED}/\textsc{SPEED+}) which corrects the non-square aspect ratio of the full image. Since the zoom can be estimated along either the $x$ or $y$-axes, we define $U_z$ as the least-squares solution satisfying $U_z \approx Z/s_x$ and $U_z \approx Z/s_y$, which yields
\begin{equation}
U_z = \frac{1}{2}\!\left(\frac{1}{s_x}+\frac{1}{s_y}\right)Z.
\label{eq:Uz_from_Z_mean}
\end{equation}
This gives us our first regression target, $U_z$. During inference, we can invert this equation to recover the absolute depth $Z$.

\paragraph{Step 2: Formulating Normalized Lateral Targets ($U_x, U_y$).}
Next, we define the targets for the lateral translation components $(X,Y)$. We start with the standard pinhole camera model, which maps a 3D point $(X,Y,Z)$ to its 2D pixel projection $(x,y)$ as
\begin{equation}
Z
\begin{bmatrix}
x \\[2pt] y \\[2pt] 1
\end{bmatrix}
=
K
\begin{bmatrix}
X \\[2pt] Y \\[2pt] Z
\end{bmatrix},
\qquad
K =
\begin{bmatrix}
f_x & 0   & c_x \\
0   & f_y & c_y \\
0   & 0   & 1
\end{bmatrix}.
\label{eq:pinhole}
\end{equation}
A key issue is that the satellite's projected 3D center on the image $(x,y)$ does not necessarily coincide with the center of its bounding box $(b_x, b_y)$. To account for this, we define our regression targets based on the offsets $(\Delta x, \Delta y) = (x-b_x, y-b_y)$. To make these targets scale-invariant and independent of the crop size, we normalize them by the crop width $w$ and height $h$. This yields
\begin{equation}
\begin{cases}
U_x := \dfrac{\Delta x}{w} = \dfrac{1}{w}\!\left(\dfrac{f_x}{Z}X + c_x - b_x\right), \\[10pt]
U_y := \dfrac{\Delta y}{h} = \dfrac{1}{h}\!\left(\dfrac{f_y}{Z}Y + c_y - b_y\right).
\end{cases}
\label{eq:UxUy_norm}
\end{equation}
The network is thus trained to regress the normalized translation vector $U=(U_x, U_y, U_z)$. At inference, once we predict $U$ and recover $Z$ from Eq.~\eqref{eq:Uz_from_Z_mean}, the lateral components $(X,Y)$ can be directly computed by inverting these relations, i.e.
\begin{equation}
\begin{cases}
X = \dfrac{1}{f_x}\Big(b_x + U_x w - c_x\Big) Z, \\[10pt]
Y = \dfrac{1}{f_y}\Big(b_y + U_y h - c_y\Big) Z.
\end{cases}
\label{eq:XY_inversion_redef}
\end{equation}

\begin{figure}[t!]
\centering
    \includegraphics[width=\linewidth]{apparent_rotation.png}
    \caption{\textbf{Apparent Rotation ($R'$).} An object's perceived orientation changes with its position in the image due to the camera perspective. The apparent rotation $R'$ is the true orientation a \emph{centered} object must have to best represent the off-center object.}
    \label{fig:apparent_rotation_definition}
\end{figure}

\subsubsection{Rotation Target: True vs.\ Apparent Orientation}

A challenge in pose regression is that perspective projection distorts an object's perceived orientation. As an object moves away from the camera's principal axis, it appears rotated, even if its true orientation $R$ remains constant.

To address this, we introduce the notion of \emph{apparent orientation} $R'$. Instead of forcing the network to learn the translation-dependent true rotation, we task the network with predicting the directly observable apparent orientation. As presented in Figure~\ref{fig:apparent_rotation_definition}, we define $R'$ as the orientation a \emph{centered} object would need to have to visually match the appearance of the translated object. Formally, we seek a corrective rotation $\Delta R$ that maps the true orientation to the apparent one,
\begin{equation}
R' = \Delta R \, R.
\label{apparent_rotation_formula}
\end{equation}
Given the camera's optical axis  $e_z=(0,0,1)^\top$ and the estimated object translation $T$ we obtain $\Delta R$ by computing the rotation axis $u$ and rotation angle $\theta$ as
\[
\vec{\boldsymbol{T}} = \frac{T}{\|T\|}, \qquad
u = \frac{\vec{\boldsymbol{T}}  \times e_z}{\|\vec{\boldsymbol{T}}  \times e_z\|}, \qquad
\theta = \arccos(\vec{\boldsymbol{T}}^\top e_z).
\]
The corrective rotation $\Delta R$ is then obtained via Rodrigues’ rotation formula as
\begin{equation}
\Delta R = I + \sin\theta [u]_\times + (1-\cos\theta)[u]_\times^2,
\end{equation}
where $[u]_\times$ is the skew-symmetric matrix associated with the vector $u$. We then recover the true orientation by inverting the corrective rotation as
\begin{equation}
R_{\text{pred}} = \Delta R^\top R'_{\text{pred}}.
\end{equation}

The network is supervised to predict only the first two columns of the apparent rotation matrix, $\hat r_1,\hat r_2 \in \mathbb{R}^3$. We construct an orthonormal basis via the Gram–Schmidt procedure, which yields
\begin{equation}
r_1 = \frac{\hat r_1}{\|\hat r_1\|}, \qquad
r_2 = \frac{\hat r_2 - (r_1^\top \hat r_2) r_1}{\|\hat r_2 - (r_1^\top \hat r_2) r_1\|}, \qquad
r_3 = r_1 \times r_2.
\end{equation}
The predicted apparent rotation is simply assembled as
\begin{equation}
R'_{\text{pred}} = \begin{bmatrix} r_1 & r_2 & r_3 \end{bmatrix}.
\end{equation}
This formulation decouples the perspective effects from the rotation prediction task, allowing the network to learn a simpler mapping from image appearance to orientation.

\section{Experiments and Results}

\subsection{Experimental Setup}

\paragraph{Data.} We train and evaluate our models on the SPEED and SPEED+ datasets. Models trained on SPEED are evaluated on SPEED, and models trained on SPEED+ are evaluated on SPEED+, treating each dataset independently.

\paragraph{Evaluation Metrics.} We use the official metrics from the ESA Pose Estimation Challenge. The translation error, $E_T$, is the Euclidean distance between the predicted and ground-truth translation vectors ($T_{pred}$ and $T_{gt}$), i.e.,
\begin{equation}
    E_T = ||T_{pred} - T_{gt}||_2.
\end{equation}
The rotation error, $E_R$, is the geodesic distance between the predicted and ground-truth orientations. For this metric, the rotations are represented as quaternions ($q$), and the error is calculated as
\begin{equation}
    E_R = 2 \cdot \arccos(|\langle q_{pred}, q_{gt} \rangle|).
\end{equation}

\paragraph{Implementation Details.} 
We initialize all ViT backbones with weights from the Facebook SWAG paper~\cite{Singh2022} pretrained on ImageNet-1K. 

For optimization, we employ the Muon optimizer~\cite{jordan2024muon}, using its default parameters but without weight decay, as we observed that enabling weight decay consistently degraded performance in our pose-regression setting.
While Muon has been primarily studied in the context of large language models~\cite{liu2025muonscalablellmtraining}, we find that it consistently outperforms AdamW in this ViT-based pose regression setting. To further improve convergence, we adopt a cosine annealing learning rate scheduler with warm restarts~\cite{loshchilov2016sgdr}, performing 5 restart periods with $T_0=10$ epochs and $T_{mult}=2$, for a total of 310 epochs on \textsc{SPEED} and 4 restarts periods for a total of 150 epochs on \textsc{SPEED+}.

It is worth noting that our extensive data augmentations help ensure the network does not overfit. The learning rate is cycled between $1 \cdot 10^{-4}$ and $1 \cdot 10^{-6}$, which helps the model escape sharp local minima. We use a batch size of 8.

For training the LW-DETR-based object detector, we use a cosine annealing learning rate scheduler with warm restarts, cycling between $5 \cdot 10^{-4}$ and $5 \cdot 10^{-5}$. We set the batch size to 16 and initialize the model with pretrained LW-DETR weights from the COCO dataset.

\paragraph{Training Losses.} 
Our model is trained with two relative losses, one for rotation and one for translation, both designed to be invariant to scale and perspective. 

The rotation loss is defined as a relative Frobenius error, i.e.,
\begin{equation}
\mathcal{L}_{\text{rot}}
= \left\| R'_{\text{pred},1:2} - R'_{\text{gt},1:2} \right\|_F^2 ,
\end{equation}
where $R'_{1:2}$ denotes the first two columns of the apparent rotation matrix.  

For translation, we supervise the normalized crop-space targets $U = (U_x, U_y, U_z)$ defined in Eq.~(6)--(7). The relative translation loss is given by
\begin{equation}
\mathcal{L}_{\text{trans}}
= \left\| U_{\text{pred}} - U_{\text{gt}} \right\|_2^2.
\end{equation}

The total loss is the simple sum of these two terms,
\begin{equation}
\mathcal{L} = \mathcal{L}_{\text{rot}} + \mathcal{L}_{\text{trans}} .
\end{equation}

These formulations yield stable gradients and significantly faster convergence compared to direct $L_2$ supervision on absolute pose parameters $(R, T)$.

\subsection{Comparison with the State of the Art}
\begin{figure*}[t]
\centering
\includegraphics[width=\textwidth]{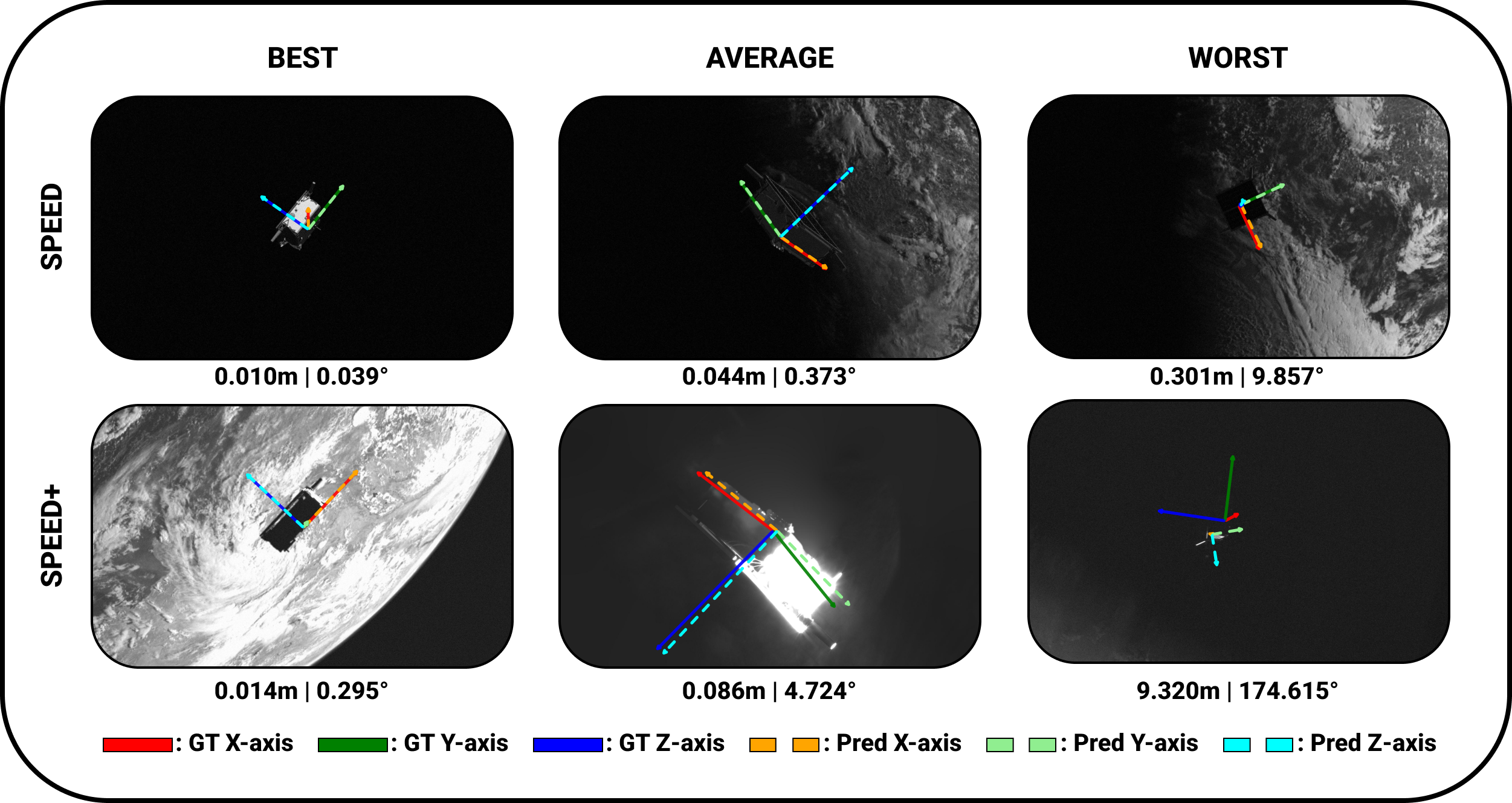}

\caption{\textbf{Qualitative Results on SPEED and SPEED+.} We visualize our best model’s performance on the test sets, showing examples of the model's best, average, and worst predictions. Ground-truth (GT) axes are in solid primary colors, while predicted (Pred) axes are shown as lighter, dashed lines. Errors below each image are reported as Translation Error [m] $\vert$ Rotation Error [deg].}
\label{fig:qualitative_results}
\end{figure*}

We compare our models against prior work on \textsc{SPEED} and \textsc{SPEED+}. On \textsc{SPEED}, our direct regression approach clearly surpasses all other non-PnP methods and matches state-of-the-art PnP pipelines, while maintaining real-time inference on edge hardware (Table~\ref{tab:sota_speed}), indicating that PnP solvers are not essential when training and testing domains align.

\begin{figure}[t!]
\centering
    \includegraphics[width=\linewidth]{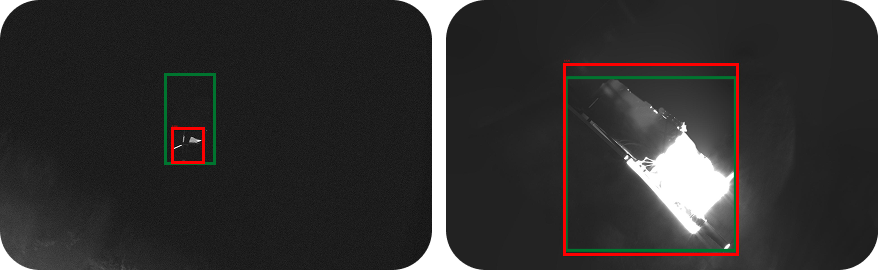}
    \caption{\textbf{Impact of bounding box quality on pose estimation.} 
    Left: a poor bounding box prediction (red) compared to the ground truth (green) leading to erroneous pose regression. 
    Right: a typical bounding box prediction (red) closely aligned with the ground truth (green) resulting in stable pose estimates.}
    \label{fig:bbox_quality_pose}
\end{figure}

On \textsc{SPEED+} (Tables~\ref{tab:sota_speedplus_lightbox}, \ref{tab:sota_speedplus_sunlamp}), the training data is synthetic as in \textsc{SPEED}, but the test data consists of real laboratory images with a large domain gap (see Fig.~\ref{fig:bbox_quality_pose}). In this setting, our method lags behind hybrid pipelines. We attribute this to two key factors:

\textbf{(1)} Leading competition teams aggressively reduced the domain gap using heavy pixel-level and style-transfer augmentations, in some cases tuned with prior knowledge of the hidden test domains. This strategy was later criticized by the organizers~\cite{SPEEDPark2023}, as it compromises competition fairness and limits reproducibility.

\textbf{(2)} Several teams reconstructed the unreleased 3D spacecraft model to synthesize large custom training datasets. While effective, this approach addresses a specific benchmark rather than the broader problem of generalizable spacecraft pose estimation.

In contrast, our pipeline relies solely on released datasets and generic augmentations, without exploiting hidden domain cues or private 3D assets. This makes our approach simple, reproducible, and generally applicable to realistic deployment scenarios, though it highlights the open challenge of achieving domain-robust direct regression.

\begin{table}[h!]
\centering
\caption{Comparison with state-of-the-art methods on \textsc{SPEED}.}
\label{tab:sota_speed}
{\small{
\begin{tabular}{@{}lccc@{}}
\toprule
\textbf{Model} & PnP? & $E_T$ [m] $\downarrow$ & $E_R$ [deg] $\downarrow$ \\ 
\midrule
\addlinespace[0.3em]
Chen et al.~\cite{Chen2019}   & Yes & 0.0320 & \textbf{0.4100} \\
\textbf{ViT-B-384/16 (Ours)} & No & \textbf{0.0221} & 0.4183 \\
\textbf{ViT-B-224/16 (Ours)} & No & 0.0231 & 0.4469 \\
Gerard et al.~\cite{Gerard2019} & Yes & 0.0730 & 0.9100 \\
E-PoseNet~\cite{Musallam2022} & No & 0.1806 & 2.3073 \\
URSONet~\cite{Proenca2019} & No  & 0.1450 & 2.4900 \\
\bottomrule
\end{tabular}
}}
\end{table}

\begin{table}[h!]
\centering
\caption{Comparison with state-of-the-art methods on the \textbf{SPEED+ Lightbox} subset. ``?" indicates unreleased information.}
\label{tab:sota_speedplus_lightbox}
{\small{
\begin{tabular}{@{}lccc@{}}
\toprule
Method & PnP? & $E_T$ [m] $\downarrow$ & $E_R$ [deg] $\downarrow$ \\ \midrule
\addlinespace[0.3em]
TangoUnchained     & Yes & 0.105 & 3.187  \\
VPU~\cite{perez2022spacecraft}                & Yes & 0.131 & 4.577  \\
SPNv2~\cite{park2024robust}              & Yes & 0.150 & 5.577  \\
lava1302~\cite{wang2023bridging}           & Yes & 0.302 & 6.665  \\
haoranhuang\_njust & Yes & 0.180 & 8.131  \\
u3s\_lab           & Yes & 0.333 & 9.694  \\
\textbf{ViT-B-384/16 (Ours)}  & No  & 0.270 & 12.261 \\
\textbf{ViT-B-224/16 (Ours)}      & No  & 0.305 & 14.668 \\
chushunhao         & ? & 0.205 & 16.378 \\
for graduate       & ? & 0.402 & 23.666 \\
Pivot SDA AI \& al.& ? & 0.409 & 23.918 \\
bbnc               & ? & 0.687 & 24.889 \\
ItTakesTwoToTango  & ? & 0.485 & 31.094 \\
\bottomrule
\end{tabular}
}}
\end{table}

\begin{table}[h!]
\centering
\caption{Comparison with state-of-the-art methods on the \textbf{SPEED+ Sunlamp} subset. ``?" indicates unreleased information.}
\label{tab:sota_speedplus_sunlamp}
{\small{
\begin{tabular}{@{}lccc@{}}
\toprule
Method & PnP? & $E_T$ [m] $\downarrow$ & $E_R$ [deg] $\downarrow$ \\ \midrule
\addlinespace[0.3em]
lava1302~\cite{wang2023bridging}            & Yes & 0.065 & 2.728  \\
VPU~\cite{perez2022spacecraft}                & Yes & 0.076 & 2.828  \\
TangoUnchained     & Yes & 0.086 & 4.299  \\
u3s\_lab           & Yes & 0.181 & 6.241  \\
haoranhuang\_njust & Yes & 0.163 & 8.406  \\
SPNv2~\cite{park2024robust}               & Yes & 0.161 & 9.788  \\
\textbf{ViT-B-384/16 (Ours)}  & No  & 0.400 & 18.277 \\
bbnc               & ? & 0.542 & 21.955 \\
for graduate       & ? & 0.455 & 22.970 \\
\textbf{ViT-B-224/16 (Ours)}      & No  & 0.460 & 28.017 \\
Pivot SDA AI \& al.& ? & 0.854 & 36.445 \\
ItTakesTwoToTango  & ? & 0.473 & 39.660 \\
chushunhao         & ? & 0.338 & 43.356 \\
\bottomrule
\end{tabular}
}}
\end{table}

\subsection{Ablation Studies and Analysis}
\label{sec:ablations}
We conduct a series of comprehensive ablation studies on the SPEED dataset to validate our key design choices. Unless otherwise specified, all experiments use our final proposed configuration on the `ViT-B-224/16` model to ensure clear and concise comparisons.

\subsubsection{Impact of Transfer Learning}
As shown in Table~\ref{tab:pretraining}, training backbones from scratch results in an order-of-magnitude drop in performance across all metrics, highlighting the importance of pre-training. This is especially true for attention-based architectures, which require large-scale supervision to learn spatial priors. Although the source task is image classification rather than pose estimation, the pre-trained weights provide a strong spatial understanding that fine-tuning can adapt to the downstream task. Without such initialization, the ViT struggles to organize attention patterns in data-limited scenarios like spacecraft pose estimation.

\begin{table}[h!]
\centering
\caption{Effect of pre-training on performance.}
\label{tab:pretraining}
{\small{
\begin{tabular}{@{}lcc@{}}
\toprule
Training & $E_T$ [m] $\downarrow$ & $E_R$ [deg] $\downarrow$ \\ \midrule
\addlinespace[0.3em]
From Scratch & 0.2391 & 5.1280 \\
Pre-trained & \textbf{0.0231} & \textbf{0.4469} \\ \bottomrule
\end{tabular}
}}
\end{table}

\subsubsection{Impact of Geometric Formulation and Cropping}
We study the effects of cropped inputs and our apparent rotation formulation. Vision Transformers operate at a fixed input size ($224 \times 224$), so directly resizing the original \textsc{SPEED}/\textsc{SPEED+} images ($1920 \times 1200$) implies a $5$–$10\times$ downscaling, causing severe loss of fine details. Cropping around the satellite alleviates this issue by ensuring that patches capture informative object regions rather than mostly background.

Within this cropped setting, our apparent rotation formulation further compensates for perspective distortions introduced when the satellite is off-center. As shown in Table~\ref{tab:geom_cropping}, both components yield cumulative improvements, with their combination achieving the best performance.

\begin{table}[h!]
\centering
\caption{Ablation of our apparent rotation formulation on both full and cropped image inputs.}
\label{tab:geom_cropping}
{\small{
\begin{tabular}{@{}llcc@{}}
\toprule
Image size & Rotation & $E_T$ [m] $\downarrow$ & $E_R$ [deg] $\downarrow$ \\ \midrule
\addlinespace[0.3em]
Full & Absolute & 0.2938 & 7.2880 \\
Full & Relative & 0.2906 & 7.3396 \\
\addlinespace[0.3em]
Cropped & Absolute & 0.0337 & 2.3778 \\
\textbf{Cropped} & \textbf{Relative} & \textbf{0.0231} & \textbf{0.4469} \\
\bottomrule
\end{tabular}
}}
\end{table}

\subsubsection{Impact of Data Augmentations}
We analyze the effect of our augmentation strategy in two stages. As shown in Table~\ref{tab:augs_impact}, spatial augmentations are crucial for \textsc{SPEED}, while pixel augmentations provide little benefit and can even be detrimental. This can be explained by the fact that both the training and test images are primarily synthetic, so pixel-level distortions do not reflect the target distribution and instead hinder learning. In contrast, bounding box perturbations (Table~\ref{tab:bbox_scale}) consistently improve robustness, with a perturbation scale of 10\% yielding the best generalization.  

\begin{table}[t]
\centering
\caption{Impact of different data augmentation on SPEED.}
\label{tab:augs_impact}
{\small{
\begin{tabular}{@{}lcc@{}}
\toprule
Augmentation Strategy & $E_T$ [m] $\downarrow$ & $E_R$ [deg] $\downarrow$ \\ \midrule
No Augmentations & 0.0616 & 1.2032 \\
Pixel Augs Only & 0.0987 & 1.4782 \\
\textbf{Spatial Augs Only} & \textbf{0.0231} & \textbf{0.4469} \\
All Augs & 0.0469 & 1.0371 \\
\bottomrule
\end{tabular}
}}
\end{table}

\begin{table}[t]
\centering
\caption{Impact of bounding box augmentations.}
\label{tab:bbox_scale}
{\small{
\begin{tabular}{@{}lcc@{}}
\toprule
BBox Perturbation (\%) & $E_T$ [m] $\downarrow$ & $E_R$ [deg] $\downarrow$ \\ \midrule
0\% & 0.0317 & 0.5787 \\
\textbf{10\%} & \textbf{0.0231} & \textbf{0.4469} \\
20\% & 0.0242 & 0.4584 \\
30\% & 0.0255 & 0.4870 \\
\bottomrule
\end{tabular}
}}
\end{table}

On the other hand, for \textsc{SPEED+}, pixel augmentations play an important role due to the domain gap between synthetic training data and real evaluation images (Table~\ref{tab:augs_impact_speedplus}). In fact, further enriching the pixel space with style-based augmentations has been shown to be highly beneficial, as demonstrated by state-of-the-art methods on \textsc{SPEED+} that leverage style transfer techniques~\cite{jackson2019style}.

\begin{table}[t]
\centering
\caption{Impact of data augmentation strategies on SPEED+ averaged over lightbox and sunlamp.}
\label{tab:augs_impact_speedplus}
{\small{
\begin{tabular}{@{}lcc@{}}
\toprule
Augmentation Strategy & $E_T$ [m] $\downarrow$ & $E_R$ [deg] $\downarrow$ \\ \midrule
No Augmentations & 1.5054 & 59.7137 \\
Pixel Augs & 0.4027 & 20.8900 \\
\textbf{Pixel Augs + Style} & \textbf{0.3506} & \textbf{18.5638} \\
\bottomrule
\end{tabular}
}}
\end{table}

\subsection{Model Compression and Deployment}
We deploy our models on a NVIDIA Jetson Orin Nano configured in 25W mode, running CUDA~12.6, TensorRT~10.7, and Python~3.10. Our focus is on the \emph{end-to-end pipeline} for spacecraft pose estimation rather than isolated inference speed. We assume that an input image is already available in memory (e.g., from an onboard camera), and process it through the following stages:  

\begin{enumerate}
    \item \textbf{Pre-processing:} Resizing with bilinear interpolation and normalization using ImageNet statistics;  
    \item \textbf{Object detection:} Running a lightweight detector to predict a single bounding box for the spacecraft;  
    \item \textbf{Cropping:} Cropping the detected region, resizing it to the network input resolution, and re-normalizing;  
    \item \textbf{Pose estimation:} Running the ViT-based pose regressor on the cropped input;  
    \item \textbf{Post-processing:} Recovering the absolute translation and rotation from the intermediate outputs.  
\end{enumerate}  

We measure the latency of each stage on the Jetson Orin Nano for different backbones (ViT-B/16 at $224\times224$ and $384\times384$ resolution) and for different inference modes (unquantized FP32 vs.\ FP16 TensorRT engines). The ViT and overhead latencies are shared in Tables~\ref{tab:deployment_latency} and \ref{tab:shared_overhead}.

\begin{table}[t]
\centering
\caption{Pose inference latency (ms) for different ViT backbones and precisions on the NVIDIA Jetson Orin Nano. Shared pre/post-processing overhead is $57.2 \pm 2.8$ ms (Table~\ref{tab:shared_overhead}).}
\label{tab:deployment_latency}
{\small{
\begin{tabular}{lccc}
\toprule
Model @ Precision & ViT Only [ms] & End-to-End [ms] \\
\midrule
ViT-B-224/16 @ FP32 & $17.0 \pm 4.2$ & $74.2 \pm 5.0$ \\
ViT-B-224/16 @ FP16 & $19.9 \pm 3.6$ & $77.1 \pm 4.6$ \\
ViT-B-384/16 @ FP32 & $49.2 \pm 7.5$ & $106.4 \pm 8.0$ \\
ViT-B-384/16 @ FP16 & $40.7 \pm 5.9$ & $97.9 \pm 6.5$ \\
\bottomrule
\end{tabular}
}}
\end{table}

\begin{table}[t]
\centering
\caption{Best averaged latency (ms) of processing steps independent of the ViT backbone. Values are mean $\pm$ std.}
\label{tab:shared_overhead}
{\small{
\begin{tabular}{cccc}
\toprule
Pre-proc & Detector & Bridge & Post-processing \\
\midrule
$13.23 \pm 0.31$ & $30.5 \pm 2.4$ & $3.7 \pm 1.3$ & $9.75 \pm 0.49$ \\
\bottomrule
\end{tabular}
}}
\end{table}

On the Jetson Orin Nano, the end-to-end latency for ViT-B-224/16 is 
about $75$\,ms per image ($\sim$13\,FPS) under strictly sequential 
execution. With non-blocking scheduling, different stages run 
concurrently (e.g., the detector processes image $i\!+\!1$ while the 
pose network handles image $i$), so throughput is limited by the 
slowest stage; this yields up to $\sim$33\,FPS with batch size $=1$. 
A key factor is detector quantization: its runtime drops from 
$71.5 \pm 3.1$\,ms in FP32 to $30.5 \pm 2.4$\,ms in FP16, a more than 
2$\times$ speedup with negligible accuracy loss.  

For the ViT backbones, FP16 shows mixed results. ViT-B-224/16 runs 
slightly faster in FP32 ($17.0 \pm 4.2$\,ms) than FP16 ($19.9 \pm 3.6$\,ms), 
as the conversion overheads outweigh the GPU acceleration for this small model. 
Larger backbones such as ViT-B-384/16 do benefit, improving from 
$49.2 \pm 7.5$\,ms (FP32) to $40.7 \pm 5.9$\,ms (FP16). Beyond latency, 
FP16 halves model size and improves hardware efficiency, making it 
preferable in most cases, except for ViT-B-224/16, where the overhead dominates.

\section{Conclusion}

We have presented FastPose-ViT, a Vision Transformer–based architecture for direct 6DoF spacecraft pose estimation. By reformulating pose regression with the concept of apparent rotation, we achieve state-of-the-art performance among non-PnP methods while maintaining a lightweight and reproducible design. A key contribution is demonstrating, for the first time, real-time ViT-based pose estimation on embedded edge hardware (up to 33 FPS on a Jetson Orin Nano), meeting the stringent requirements of spaceborne deployment. Our ablations validate the importance of pre-training, cropping, and geometric targets in this setting. While challenges remain under large domain gaps and bounding-box sensitivity, training on datasets that better reflect real-world spacecraft imagery could significantly reduce this gap and improve generalization.

{\small
\bibliographystyle{ieeenat_fullname}
\bibliography{main}
}

\newpage
\subsection{Derivation for Spatial Augmentation}
In the main paper, we state that for an in-plane rotation matrix $M$ and camera intrinsic matrix $K$, the term $K^{-1}MK$ simplifies to a pure 3D rotation matrix $R_z(\theta)$. Here we provide the proof.

The intrinsic matrix $K$ and its inverse $K^{-1}$ are given by:
\begin{equation}
K=\begin{pmatrix}f_{x}&0&W/2\\ 0&f_{y}&H/2\\ 0&0&1\end{pmatrix}, K^{-1}=\begin{pmatrix}\frac{1}{f_x}&0&-\frac{W}{2f_x}\\ 0&\frac{1}{f_y}&-\frac{H}{2f_y}\\ 0&0&1\end{pmatrix}
\end{equation}
The in-plane rotation matrix $M(\theta)$ which rotates an image by angle $\theta$ around its center $(W/2, H/2)$ is:
\begin{equation}
M(\theta)=\begin{pmatrix}\cos\theta&-\sin\theta&\frac{W}{2}(1-\cos\theta)+\frac{H}{2}\sin\theta\\ \sin\theta&\cos\theta&\frac{H}{2}(1-\cos\theta)-\frac{W}{2}\sin\theta\\ 0&0&1\end{pmatrix}
\end{equation}
By performing the matrix multiplication for $K^{-1}M(\theta)$ and then $K^{-1}M(\theta)K$, and assuming $f_x = f_y$ (a common case), the expression simplifies significantly.
The final result of the multiplication $K^{-1}M(\theta)K$ is:
\begin{equation}
K^{-1}M(\theta)K = \begin{pmatrix}\cos\theta&-\sin\theta&0\\ \sin\theta&\cos\theta&0\\ 0&0&1\end{pmatrix} = R_z(\theta)
\end{equation}
This is the standard rotation matrix for a counter-clockwise rotation of angle $\theta$ about the z-axis. It is an orthogonal matrix with a determinant of 1, confirming it is a valid rotation matrix in $SO(3)$.
\subsection{ViT Backbone Performance Trade-Off }

In Table~\ref{tab:vit_backends} we summarize the performance of various ViT backbones. As expected, larger models with higher resolution and smaller patch sizes generally perform better but require more floating point operations. We focus our analysis on the ViT-B-224/16 and ViT-B-384/16 due to their favorable trade-off between performance and latency.

\begin{table}[h!]
\centering
\caption{Performance of various ViT backbones on SPEED.}
\label{tab:vit_backends}
\resizebox{\linewidth}{!}{
\begin{tabular}{@{}lcccc@{}}
\toprule
ViT Backend & GFLOPS  & Params (M)  & $E_T$ [m] $\downarrow$ & $E_R$ [deg] $\downarrow$ \\ \midrule
ViT-B-224/16 & 17.6 & 86.6 & 0.0231 & 0.4469 \\
ViT-B-384/16 & 55.5 & 86.9 & 0.0221 & 0.4183 \\
ViT-B-224/32 & 4.4 & 88.2 & 0.0425 & 0.7907 \\
\addlinespace
ViT-L-224/16 & 61.55  & 304.3 & 0.0218 & 0.4068 \\
ViT-L-512/16 & 361.99 & 305.2 & \textbf{0.0182} & \textbf{0.3380} \\
ViT-L-224/32 & 15.38  & 306.5 & 0.0437 & 0.8079 \\
\addlinespace
ViT-H-224/14 & 167.29 & 632.0 & 0.0211 & 0.3896 \\
ViT-H-518/14 & 1016.7 & 633.5 & 0.0188 & 0.3438 \\
\bottomrule
\end{tabular}
}
\end{table}

\end{document}